\documentclass[runningheads]{llncs}
\usepackage[T1]{fontenc}
\usepackage{graphicx,verbatim}
\usepackage{adjustbox}
\usepackage{adjustbox}
\usepackage{amsmath}
\usepackage{amsfonts}
\usepackage{array}

\usepackage{booktabs}
\usepackage{graphicx}
\usepackage{makecell}
\usepackage{multirow}
\usepackage{ragged2e}
\usepackage{siunitx}
\usepackage[colorinlistoftodos]{todonotes}
\usepackage{xcolor}
\usepackage{arydshln}
\begin{document}
\title{PathMoE: Interpretable Multimodal Interaction Experts for Pediatric Brain Tumor Classification}
\titlerunning{PathMoE}

%




\author{
Jian Yu\inst{1,*} \and
Joakim Nguyen\inst{1,*} \and
Jinrui Fang\inst{1} \and
Awais Naeem\inst{1} \and
Zeyuan Cao\inst{1} \and
Sanjay Krishnan\inst{2} \and
Nicholas Konz\inst{3} \and
Tianlong Chen\inst{3} \and
Chandra Krishnan\inst{2} \and
Hairong Wang\inst{1} \and
Edward Castillo\inst{1} \and
Ying Ding\inst{1,\dagger} \and
Ankita Shukla\inst{4,\dagger}
\thanks{* Equal contribution. $\dagger$ Corresponding authors.}
}
\authorrunning{Yu, Nguyen et al.}
\institute{
University of Texas, Austin, TX, USA \and
Dell Children's Medical Center, Austin, TX, USA \and
University of North Carolina at Chapel Hill, NC, USA \and
University of Nevada, Reno, NV, USA \\
\email{\{jian.yu, jhn001\}@utexas.edu} \\
\email{ying.ding@utexas.edu, ankitas@unr.edu}
}
\maketitle         
%
\begin{abstract}

Accurate classification of pediatric central nervous system tumors remains challenging due to histological complexity and limited training data. While pathology foundation models have advanced whole-slide image (WSI) analysis, they often fail to leverage the rich, complementary information found in clinical text and tissue microarchitecture. To this end, we propose \textbf{PathMoE}, an interpretable multimodal framework that integrates H\&E slides, pathology reports, and nuclei-level cell graphs via an interaction-aware mixture-of-experts architecture built on state-of-the-art foundation models for each modality. By training specialized experts to capture modality uniqueness, redundancy, and synergy, PathMoE employs an input-dependent gating mechanism that dynamically weights these interactions, providing sample-level interpretability.
We evaluate our framework on two dataset-specific classification tasks on an internal pediatric brain tumor dataset (PBT) and external TCGA datasets. PathMoE improves macro-F1 from 0.762 to 0.799 (+0.037) on PBT when integrating WSI, text, and graph modalities; on TCGA, augmenting WSI with graph knowledge improves macro-F1 from 0.668 to 0.709 (+0.041). These results demonstrate significant performance gains over state-of-the-art image-only baselines while revealing the specific modality interactions driving individual predictions. This interpretability is particularly critical for rare tumor subtypes, where transparent model reasoning is essential for clinical trust and diagnostic validation.


\end{abstract}
\section{Introduction}
Pediatric brain tumors are the leading cause of cancer-related mortality in children, making precise classification critical for treatment planning and prognosis.
However, histopathological diagnosis remains challenging due to pronounced histologic heterogeneity, limited training data, and the prevalence of rare tumor subtypes \cite{ostrom2020cbtrus,louis2021who}.
In clinical practice, diagnosis is informed by multiple complementary sources, including hematoxylin and eosin (H\&E) whole-slide images (WSIs), pathology reports, and tissue microarchitecture \cite{louis2021who,lu2023pathologycopilot}, yet current computational methods remain largely image-only and fail to leverage rich, contextual multimodal evidence, resulting in limited diagnostic utility \cite{ilse2018attention,lu2021clam,ding2025multimodal}.

Early advances in WSI analysis were largely driven by weakly supervised learning under the multiple instance learning (MIL) framework, where attention-based aggregation enables slide-level prediction from patch features \cite{ilse2018attention}.
Subsequent work improves robustness and discriminative capacity through clustering-based MIL \cite{lu2021clam}, transformer-based modeling of patch dependencies \cite{shao2021transmil}, and structured state-space formulations \cite{fillioux2023structured}.
Concurrently, large-scale pretrained visual encoders, including general-purpose deep backbones \cite{he2015deepresiduallearningimage} and pathology-specific foundation models such as UNI \cite{chen2024uni}, have substantially enhanced representation quality.
Despite these advances, image-only pipelines are still insufficient for difficult pediatric cases, where subtle morphologic differences and overlapping visual patterns commonly confound analysis.

To address these limitations, recent work explores integrating WSIs with textual information from pathology reports or captions.
Foundation models like CONCH \cite{lu2024avisionlanguage} and TITAN \cite{ding2025multimodal} align these modalities via large-scale pretraining, while generative systems like PathChat \cite{lu2023pathologycopilot} and SlideChat \cite{chenSlideChatLargeVisionLanguage2024} support interactive workflows.
Broader efforts extend multimodal modeling to oncology-focused foundations \cite{xiang2025vision}, molecularly informed representations \cite{vaidya2025molecular}, and general medical vision-language models \cite{chen2024towards}. However, these general-purpose models often fail to capture the subtle morphologic nuances and rare subtypes characteristic of pediatric brain tumors, necessitating the integration of explicit domain knowledge to resolve complex diagnostic cues.

One important source of such domain knowledge is tissue microarchitecture, which reflects cellular organization and spatial interactions within the tumor microenvironment.
Graph-based representations model this structure by constructing nuclei graphs from instance segmentations \cite{graham2019hovernetsimultaneoussegmentationclassification,10.1007/978-3-030-23937-4_2}, enabling multi-scale representations of tissue structure via frameworks such as HistoCartography \cite{jaume2021} and others \cite{pati2021}. However, current methods often treat graph features as isolated inputs, failing to explicitly model the interactions between structured tissue architecture and complementary modalities.
Consequently, it remains challenging to understand how information from different modalities can be jointly leveraged for multimodal prediction.

We introduce \textbf{PathMoE}, a knowledge-guided multimodal framework for pediatric brain tumor classification.
PathMoE integrates nuclei-level cell graphs as structured domain knowledge to guide the fusion of WSIs and pathology reports within an interaction-aware mixture-of-experts (MoE) architecture. An input-dependent gating mechanism enables interpretable assessment of modality contributions for each prediction. Experiments demonstrate consistent gains across datasets: on PBT, macro-F1 improves from 0.762 to 0.799 (+0.037), and on the TCGA subset, macro-F1 improves from 0.668 to 0.709 (+0.041). \textbf{Our key contributions include:} \textbf{(1)} A multimodal framework incorporating nuclei-level cell graphs as structured domain knowledge to guide learning from WSIs and pathology report text; \textbf{(2)} an interpretable MoE architecture that quantifies sample-level modality contributions and cross-modal interactions; and \textbf{(3)} clinically-grounded error analysis conducted by a neuropathologist, providing expert validation of model predictions and failure cases.

\begin{figure*}[t!]
    \centering
    \includegraphics[width=\linewidth]{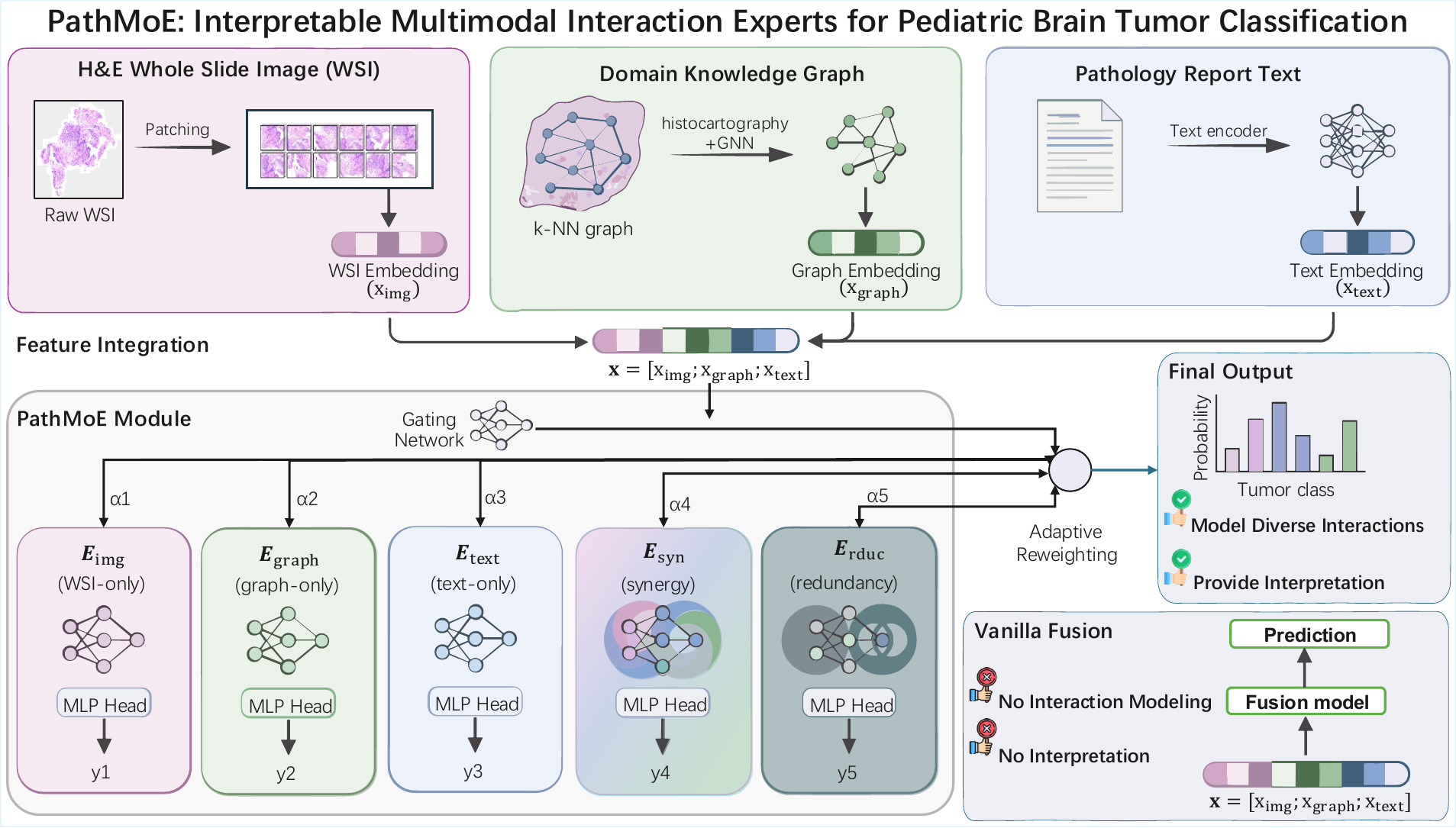}
    \caption{Overview of \textbf{PathMoE}. H\&E WSIs, pathology reports, and nuclei graphs are encoded and fused via an interaction-aware mixture-of-experts module. An input-dependent gating network computes sample-specific weights to combine expert predictions into the final tumor classification. A vanilla fusion method is also depicted.}
    \label{fig:Figure 1}
\end{figure*}



\section{Methods}
An overview of \textbf{PathMoE} is illustrated in Fig.~\ref{fig:Figure 1}. Each pediatric case is represented by three modalities: H\&E whole-slide images ($x_{\text{img}}$), nuclei-level tissue microarchitecture ($x_{\text{graph}}$), and pathology reports ($x_{\text{text}}$). Embeddings for each modality are processed by the PathMoE module, an interaction-aware mixture-of-experts (MoE) architecture comprising five experts. Three unimodal experts ($E_{\text{img}}, E_{\text{graph}}, E_{\text{text}}$) capture modality-specific features, while two interaction experts model synergy ($E_{\text{syn}}$) and redundancy ($E_{\text{rduc}}$). Unlike vanilla fusion, which simply concatenates features without capturing cross-modal relationships, our gating network dynamically weighs these experts to model complex inter-modality interactions, thereby enhancing diagnostic performance and providing interpretable insight for tumor subtype predictions.

\subsection{Multimodal Feature Extraction}
Each modality $m \in \{\text{img, graph, text}\}$ is encoded into a slide-level representation $x_m \in \mathbb{R}^{d_m}$ via the following methods, which is then projected to a set of fusion tokens $X_m \in \mathbb{R}^{P \times d}$ ($P=16$ by default).

\noindent \textbf{WSI Visual Feature Extraction.}
WSIs are tiled into non-overlapping patches $\{p_i\}_{i=1}^{N}$, with patch embeddings $h_i \in \mathbb{R}^{d_0}$ extracted via a pretrained foundation model (UNIv2~\cite{chen2024uni}). WSI preprocessing is performed with TRIDENT at 20$\times$ magnification using 256$\times$256 patches; tissue regions are detected via HEST segmentation.~\cite{zhang2025standardizing,vaidya2025molecular,jaume2024hest}. To obtain the slide-level representation $x_{\text{img}}$, we utilize gated attention MIL~\cite{ilse2018attention} to aggregate patch features using learned (softmax) attention weights $a_i$:
$x_{\text{img}} = \sum_{i=1}^{N} a_i \phi_{\text{img}}(h_i)$
where $\phi_{\text{img}}$ is a projection layer. This representation is further transformed into the final fusion tokens $X_{\text{img}}$.

\noindent \textbf{Textual Feature Extraction.}
Pathology report microscopic descriptions are encoded using a domain-specific text encoder foundation model (TITAN \cite{ding2025multimodal} by default). Following a similar paradigm to the visual branch, the encoder $f_{\text{text}}$ processes the report text to yield a global representation $x_{\text{text}}$ via its final hidden state or an attention-pooling layer (depending on the encoder), which is subsequently projected to $X_{\text{text}}$.


\noindent \textbf{Cell Graph Construction and Graph Embeddings.}
To capture tissue microarchitecture, we construct nuclei graphs from 4$\times$ downscaled WSIs using \texttt{histocartography}~\cite{jaume2021}. Nuclei are detected via HoverNet~\cite{graham2019hovernetsimultaneoussegmentationclassification} and encoded via ResNet34 features extracted from the a 224×224-pixel region centered on each nucleus. We construct a $k$-NN graph ($k=5$) where nodes represent nuclei and edges represent spatial proximity. Node embeddings are updated via GraphSAGE:
\[
h_v^{(\ell+1)}=\sigma\left(W_1^{(\ell)}h_v^{(\ell)} + W_2^{(\ell)}\cdot \mathrm{mean}_{u\in\mathcal{N}(v)} h_u^{(\ell)}\right),
\]
and final graph-level features $x_{\text{graph}}$ are obtained via attention MIL pooling over node embeddings and projected to tokens $X_{\text{graph}}$.



\subsection{Interaction-aware Mixture of Experts}
The modality tokens $\mathbf{X} = \{X_{\text{img}}, X_{\text{text}}, X_{\text{graph}}\}$ are integrated via an Interaction-aware Mixture-of-Experts (I$^2$MoE) framework~\cite{xin2025i2moe}. Unlike vanilla fusion, I$^2$MoE explicitly models cross-modal relationships through $K=5$ specialized experts: three unimodal uniqueness experts ($\mathcal{E}_{\text{img}}, \mathcal{E}_{\text{text}}, \mathcal{E}_{\text{graph}}$), a synergy expert ($\mathcal{E}_{\text{syn}}$), and a redundancy expert ($\mathcal{E}_{\text{rduc}}$).


\noindent\textbf{Expert Mechanism.} Each expert $k$ computes a clean output $y_k^{(0)} = \mathcal{E}_k(\mathbf{X})$. To isolate specific interactions, we utilize a perturbation strategy where $M$ additional outputs $y_k^{(r)} = \mathcal{E}_k(R_r(\mathbf{X}))$ are computed by replacing modality $r$ with a random tensor. This allows the gating network to distinguish information specific to one modality from information shared or synergistic across modalities.


\noindent \textbf{Adaptive Reweighting and Loss.} To achieve sample-specialized fusion, we employ a gating MLP, $g$, which takes the concatenated global representations $\mathbf{x} = [x_{\text{img}}, x_{\text{text}}, x_{\text{graph}}]$ as input. The gating network computes the expert weights $\alpha \in \mathbb{R}^K$ as $\alpha = \text{softmax}(g(\mathbf{x}))$,
where $\alpha_k$ represents the relative contribution of expert $k$ to the final prediction. The final classification logits are then computed as the weighted sum of expert outputs, $y = \sum_{k=1}^{K} \alpha_k y_k^{(0)}$. This mechanism ensures that the model dynamically prioritizes specific modalities or interactions based on the unique diagnostic evidence present in each sample.

The model is optimized using $\mathcal{L} = \mathcal{L}_{\text{cls}} + \lambda_{\text{int}}\mathcal{L}_{\text{int}}$, where the interaction loss $\mathcal{L}_{\text{int}}$ regularizes the unimodal, synergistic, and redundant components to ensure expert specialization and gating interpretability \cite{xin2025i2moe} ($\lambda_{\text{int}}=0.1$ by default), and $\mathcal{L}_{\text{cls}}$ is a standard classification loss.
\section{Experimental Setup}

\textbf{Datasets.} We primarily utilize a de-identified WSI cohort collected from our institution under IRB approval, comprising 253 diagnostic H\&E WSIs from 196 patients (age $\leq$ 29 years), labeled the \textbf{PBT Dataset}. We categorized 199 cases into four diagnostic classes: \emph{Low-grade CNS tumor}, \emph{High-grade CNS tumor}, \emph{Non-glial tumor}, and \emph{Ependymoma}. Patient-level splits were enforced to prevent data leakage.
For external evaluation, we included diagnostic H\&E WSIs from TCGA-GBM and TCGA-LGG accessed via the Genomic Data Commons (GDC)\cite{weinstein2013tcga}, forming the \textbf{TCGA Dataset} (age $\leq$ 29 years). The TCGA dataset comprises 208 WSIs from 67 patients, its diagnostic classes include \emph{Astrocytoma}, \emph{Glioblastoma}, and \emph{Oligodendroglioma}, as recorded in the clinical metadata. All models were evaluated via 10-fold cross-validation each with 80\%/10\%/10\% randomized patient-level splits.


\noindent \textbf{Baseline Models.} We establish a strong unimodal baseline using four competitive Multiple Instance Learning (MIL) models: CLAM \cite{lu2021clam}, TransMIL \cite{shao2021transmil}, S4MIL \cite{fillioux2023structured}, and MambaMIL \cite{yang2024mambamil}, representing attention, transformer, and state-space aggregation paradigms. To ensure a controlled comparison, all baselines utilize the same UNIv2 \cite{chen2024uni} patch embeddings.

\noindent \textbf{PathMoE Implementation.} We evaluate our interaction-aware mechanism across two common multimodal fusion techniques: (1) the Early Fusion (EF) backbone, which uses a transformer to jointly encode concatenated modality features, and (2) the SwitchGate (SG)~\cite{JMLR:v23:21-0998} backbone, which adaptively re-weights modality contributions via a gating mechanism. 


\noindent \textbf{Evaluation Metrics.} We report mean macro-F1 as the primary metric, alongside macro and per-class precision and recall. All metrics are derived from test-set predictions using checkpoints selected via the validation set. Finally, all experiments were conducted on a single NVIDIA GH200 GPU (96GB HBM).

\section{Results}
\subsubsection{Image-Only Baselines.}
We first compare WSI-only baselines of CLAM, MambaMIL, S4MIL, TransMIL, and WSI-only versions of the EF and SG backbones, denoted respectively as EF$_W$ and SG$_W$.
On the PBT dataset (Table \ref{tab:results_multiclass_pbt}), CLAM achieves the highest image-only macro-F1 (0.764), followed closely by EF$_W$ (0.762). SG$_W$ performs poorly (0.618), indicating that gating mechanisms require multi-modality inputs to function effectively. Similar trends emerge on TCGA (Table~\ref{tab:results_multiclass_tcga}), where EF$_W$ (0.668) leads over CLAM (0.651) and other baselines including SG$_W$. Overall, while WSI features capture strong morphological cues, performance remains inconsistent across lineages, with Glioblastoma proving easier to classify than more challenging subtypes like Oligodendroglioma.

\begin{table*}[htbp]
\centering
\caption{\textbf{Results on PBT:} per-class precision (P), recall (R), and F1 across four classes and macro average. Abbrev.: Epend.=Ependymoma; LG-CNS=Low-grade CNS tumor; HG-CNS=High-grade CNS tumor; NGT=Non-glial tumor; PM=PathMoE; EF=Early Fusion; SG=SwitchGate.}
\label{tab:results_multiclass_pbt}

\begingroup
\setlength{\tabcolsep}{1.2pt} 
\renewcommand{\arraystretch}{1.06}
\setlength{\heavyrulewidth}{1.2pt}
\setlength{\lightrulewidth}{0.6pt}
\setlength{\cmidrulewidth}{0.45pt}

\fontsize{8pt}{9.6pt}\selectfont

\begin{adjustbox}{center}
\begin{tabular}{@{}>{\raggedright\arraybackslash}p{1.7cm} | *{4}{ccc| } ccc @{}}
\toprule
& \multicolumn{3}{c}{Epend.}
& \multicolumn{3}{c}{LG-CNS}
& \multicolumn{3}{c}{HG-CNS}
& \multicolumn{3}{c}{NGT}
& \multicolumn{3}{c}{Macro} \\
\cmidrule(lr){2-4}\cmidrule(lr){5-7}\cmidrule(lr){8-10}\cmidrule(lr){11-13}\cmidrule(lr){14-16}
Method
& P & R & F1
& P & R & F1
& P & R & F1
& P & R & F1
& P & R & F1 \\
\midrule

CLAM
& \textbf{.600} & \textbf{.600} & \textbf{.600}
& .753 & .824 & .780
& \textbf{.864} & .847 & .852
& .897 & .800 & .824
& .778 & .768 & \underline{.764} \\

MambaMIL
& \underline{.533} & \textbf{.600} & \underline{.550}
& .751 & .838 & .782
& \underline{.858} & .868 & \underline{.857}
& .897 & .675 & .741
& .760 & .745 & .733 \\

S4MIL
& .500 & \underline{.500} & .500
& .751 & .824 & .774
& .853 & \underline{.880} & \textbf{.861}
& .817 & .650 & .712
& .730 & .713 & .712 \\

TransMIL
& .450 & \underline{.500} & .467
& .782 & .805 & .765
& .824 & .869 & .829
& .930 & .625 & .702
& .746 & .700 & .691 \\

EF$_{\mathrm{W}}$
& \textbf{.600} & \textbf{.600} & \textbf{.600}
& .852 & .901 & .873
& .790 & .807 & .787
& .875 & .750 & .790
& \underline{.779} & .765 & .762 \\

SG$_{\mathrm{W}}$
& .100 & .100 & .100
& .857 & .858 & .848
& .711 & \textbf{.890} & .780
& .908 & .650 & .746
& .644 & .625 & .618 \\

PM-EF$_{\mathrm{WT}}$
& .500 & \underline{.500} & .500
& \underline{.903} & .878 & \underline{.881}
& .784 & .857 & .803
& .922 & \textbf{.850} & \textbf{.868}
& .777 & \underline{.771} & .763 \\

PM-EF$_{\mathrm{WG}}$
& .350 & .400 & .367
& .850 & .901 & .872
& .759 & .857 & .796
& .883 & .650 & .728
& .711 & .702 & .691 \\

PM-EF$_{\mathrm{WTG}}$
& \textbf{.600} & \textbf{.600} & \textbf{.600}
& \textbf{.911} & \textbf{.923} & \textbf{.914}
& .802 & .857 & .821
& .935 & \underline{.825} & \underline{.862}
& \textbf{.812} & \textbf{.801} & \textbf{.799} \\

PM-SG$_{\mathrm{WT}}$
& .300 & .300 & .300
& .866 & .880 & .866
& .752 & \textbf{.890} & .807
& \underline{.975} & .750 & .837
& .723 & .705 & .703 \\

PM-SG$_{\mathrm{WG}}$
& .400 & .400 & .400
& .865 & \underline{.902} & .878
& .746 & \textbf{.890} & .800
& \textbf{.980} & .650 & .753
& .748 & .711 & .708 \\

PM-SG$_{\mathrm{WTG}}$
& .500 & \underline{.500} & .500
& .878 & .869 & .869
& .755 & .857 & .789
& .925 & .750 & .811
& .764 & .744 & .742 \\

\bottomrule
\end{tabular}
\end{adjustbox}

\endgroup
\end{table*}

\noindent \textbf{Integrating Multimodal Information and Domain Knowledge.}
Integrating pathology reports ($T$) and nuclei graphs ($G$) consistently enhances performance. Our proposed PathMoE-EF$_{WTG}$ achieves the best overall performance on PBT (macro-F1: 0.799), surpassing the strongest image-only model (CLAM, 0.764), as well as our model's strongest two-modality variants (PathMoE-EF$_{WT}$, 0.763 and PathMoE-EF$_{WG}$, 0.691). Adding report text $T$ (comparing PathMoE-$EF_{WT}$ vs. EF$_W$) yields notable class-specific gains, particularly for Non-glial (F1: 0.868 vs. 0.790) and Low-grade CNS tumors (F1: 0.881 vs. 0.873), indicating that reports provide complementary information beyond image appearance, though these are partially offset by a drop for Ependymoma.
\begin{table*}[htbp]
\centering
\caption{\textbf{Results on TCGA:} per-class precision (P), recall (R), and F1 across three classes and macro average. Abbrev.: Oligo.=Oligodendroglioma; Astro.=Astrocytoma; GBM=Glioblastoma.}
\label{tab:results_multiclass_tcga}

\begingroup
\setlength{\tabcolsep}{1.2pt}
\renewcommand{\arraystretch}{1.06}
\setlength{\heavyrulewidth}{1.2pt}
\setlength{\lightrulewidth}{0.6pt}
\setlength{\cmidrulewidth}{0.45pt}

\fontsize{8pt}{9.6pt}\selectfont

\begin{adjustbox}{center}
\begin{tabular}{@{}>{\raggedright\arraybackslash}p{2.1cm} | *{3}{ccc| } ccc @{}}
\toprule
& \multicolumn{3}{c}{Oligo.}
& \multicolumn{3}{c}{Astro.}
& \multicolumn{3}{c}{GBM}
& \multicolumn{3}{c}{Macro} \\
\cmidrule(lr){2-4}\cmidrule(lr){5-7}\cmidrule(lr){8-10}\cmidrule(lr){11-13}
Method
& P & R & F1
& P & R & F1
& P & R & F1
& P & R & F1 \\
\midrule

CLAM~\cite{lu2021clam}
& .652 & .534 & .528
& .635 & .602 & .576
& \textbf{.860} & .857 & \underline{.849}
& \underline{.716} & .664 & .651 \\

MambaMIL~\cite{yang2024mambamil}
& .634 & \textbf{.678} & \underline{.596}
& .588 & .556 & .544
& .807 & .769 & .777
& .677 & .668 & .639 \\

S4MIL~\cite{fillioux2023structured}
& \underline{.675} & .407 & .458
& .594 & \textbf{.667} & \underline{.611}
& .803 & .872 & .830
& .691 & .649 & .633 \\

TransMIL~\cite{shao2021transmil}
& .526 & .509 & .496
& .550 & .558 & .533
& .812 & .857 & .821
& .629 & .641 & .617 \\

EF$_{\mathrm{W}}$~\cite{xin2025i2moe}
& .592 & \underline{.637} & .590
& \underline{.659} & .565 & .586
& \underline{.826} & .843 & .827
& .693 & \underline{.682} & \underline{.668} \\

SG$_{\mathrm{W}}$~\cite{JMLR:v23:21-0998}
& .557 & .515 & .499
& .654 & .532 & .527
& .805 & \underline{.893} & .840
& .672 & .647 & .622 \\

PM-EF$_{\mathrm{WG}}$
& \textbf{.779} & .580 & \textbf{.629}
& \textbf{.699} & \underline{.655} & \textbf{.665}
& .798 & \underline{.893} & .835
& \textbf{.759} & \textbf{.709} & \textbf{.709} \\

PM-SG$_{\mathrm{WG}}$
& .569 & .489 & .477
& .645 & .556 & .545
& .820 & \textbf{.899} & \textbf{.853}
& .678 & .648 & .625 \\

\bottomrule
\end{tabular}
\end{adjustbox}

\endgroup
\end{table*}
The graph modality ($G$) also provides critical, non-redundant structural priors; adding $G$ to the $WT$ setting yields a +0.036 to +0.039 macro-F1 boost across both fusion backbones. Notably, SG$_{WG}$ (0.708) significantly improves over SG$_W$ (0.618), and greatly improves performance on the Ependymoma class. For the switch-gating branch, PathMoE-SG$_{WT}$ improves macro-F1 from 0.618 to 0.703 (+0.085) with similar gains in precision and recall, suggesting that text features stabilize representations when image-only gating is weak. The importance of $G$ is even more pronounced on TCGA  (Table~\ref{tab:results_multiclass_tcga}), where text is not usable due to document noise, and the strongest improvements come from incorporating graph/domain knowledge. In particular, EF$_{WG}$ achieves a macro-F1 of 0.709, improving over EF$_W$ (0.668, second-best) and all other image-only baselines. This gain is driven by substantial improvements on the more challenging classes of Oligodendroglioma and Astrocytoma, while maintaining strong performance on Glioblastoma. These results demonstrate that structured graph/domain information is a critical modality for improving robustness and accuracy when report text is unavailable or unreliable.

\newcolumntype{L}[1]{>{\raggedright\arraybackslash}m{#1}}
\newcolumntype{C}[1]{>{\centering\arraybackslash}m{#1}}
\newcolumntype{J}{>{\justifying\setlength{\parindent}{0pt}\arraybackslash}m{3.8cm}}

\newcommand{\vcenterimg}[2][1.4cm]{%
  \raisebox{-0.1\height}{\includegraphics[width=#1]{#2}}}

\begin{table*}[htbp]
\renewcommand{\arraystretch}{1}
\setlength{\tabcolsep}{1.5pt}
\fontsize{8}{8}\selectfont
\centering
\setlength{\dashlinedash}{0.5pt}
\setlength{\dashlinegap}{2pt}

\caption{Qualitative error analysis of the baseline and PathMoE (PM) across image–text (WT) and image–text–graph (WTG) settings. Each case shows the ground-truth label, model predictions, modality interaction weights ($w_W$, $w_T$, $w_G$), and a brief clinical rationale for the correct diagnosis. Nodes and edges (light blue) often aggregate in regions of cellular density learned by the GNN.}

\begin{tabular}{L{0.2cm} C{1.4cm} C{1cm} C{1.3cm} C{1.3cm} C{0.8cm} C{0.8cm} C{0.8cm} J}
    \toprule
      & 
    {\textbf{Case}}
    &
    {\textbf{Label}} &
    {\textbf{Baseline}} &
    $\textbf{PM}_\textbf{WTG}$ &
    $w_W$ & $w_T$ & $w_G$ &
    \textbf{Clinical Rationale} \\
    \midrule

    \rotatebox{90}{\texttt{Case\_1}} &
    \vcenterimg{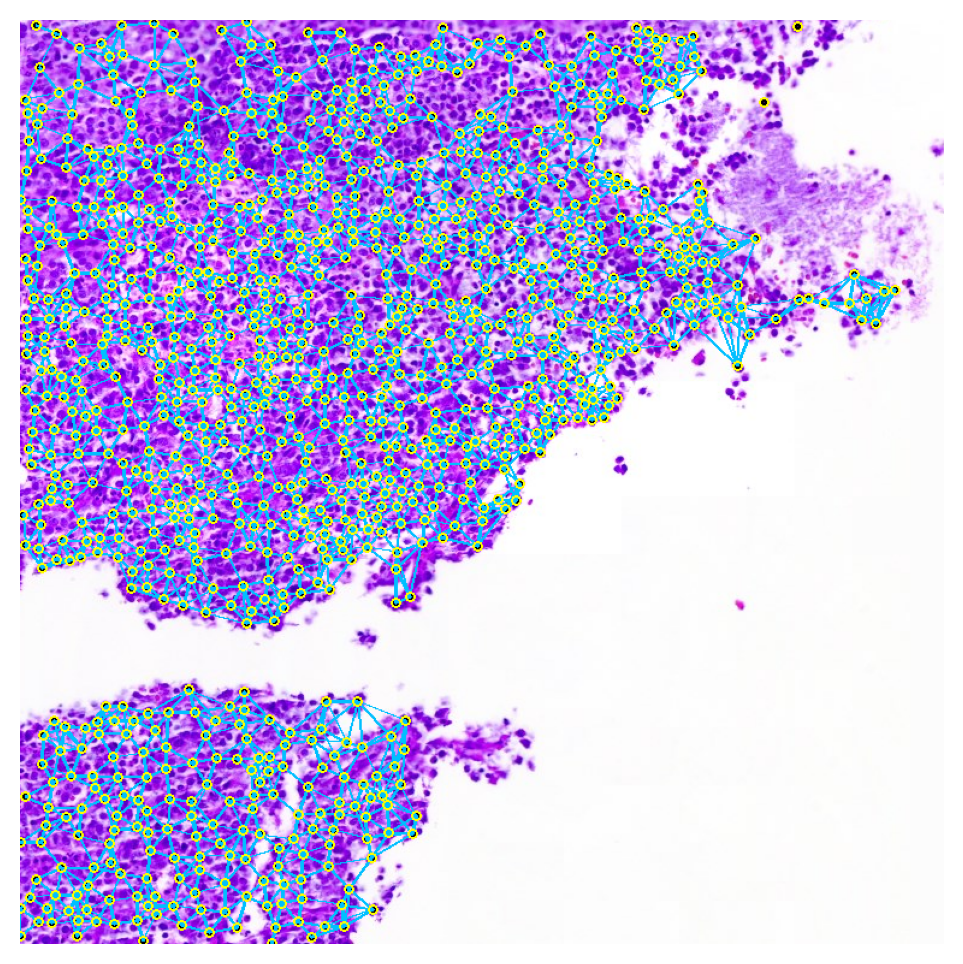} &
    \textbf{HG-CNS} &
    NGT &
    \textcolor{green!50!black}{\textbf{HG-CNS}} &
    0.232 & 0.167 & 0.173 &
    Polyphenotypic tumor mimicking non-glial types; rhabdoid cells and monotonous morphology favor HG despite overlap. \\
    \hline
    
    \rotatebox{90}{\texttt{Case\_2}} &
    \vcenterimg{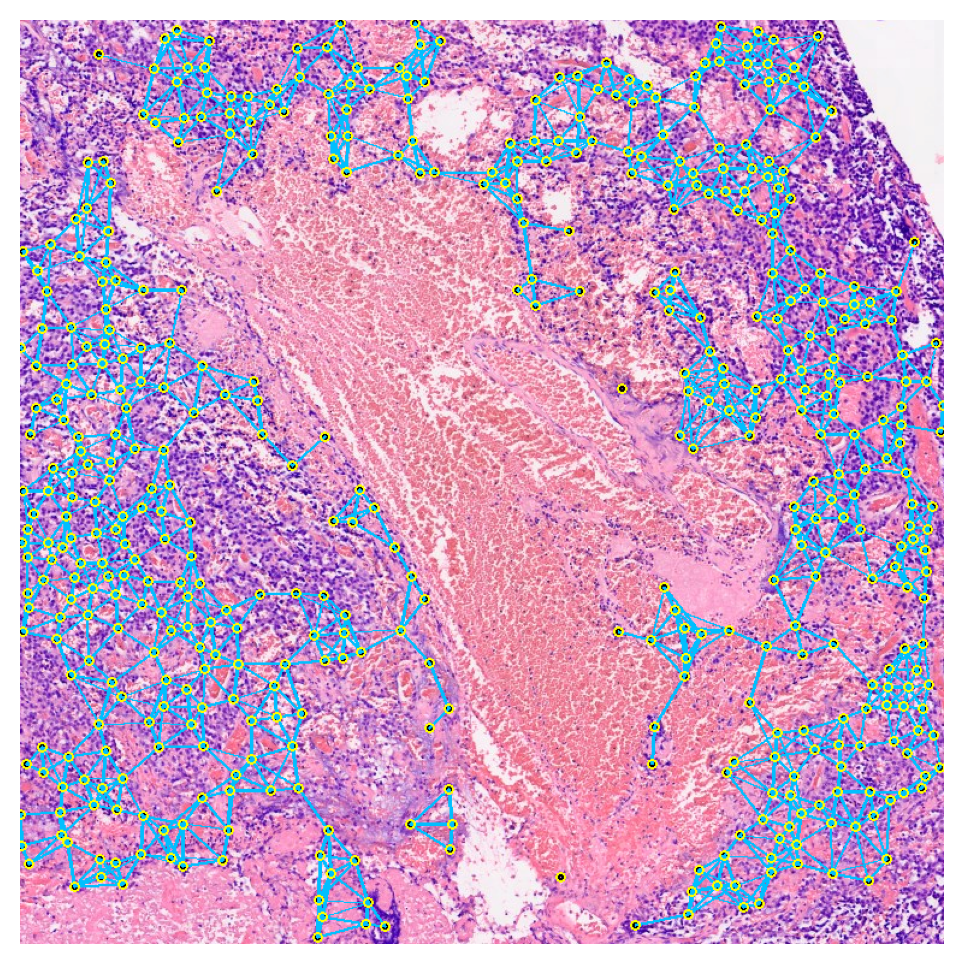} &
    \textbf{NGT} &
    HG-CNS &
    \textcolor{green!50!black}{\textbf{NGT}} &
    0.208 & 0.184 & 0.205 &
    High cell density and hemorrhage mimic HGs; lacks nuclear atypia/mitoses and shows pituitary morphology. \\
    \hline

    \rotatebox{90}{\texttt{Case\_3}} &
    \vcenterimg{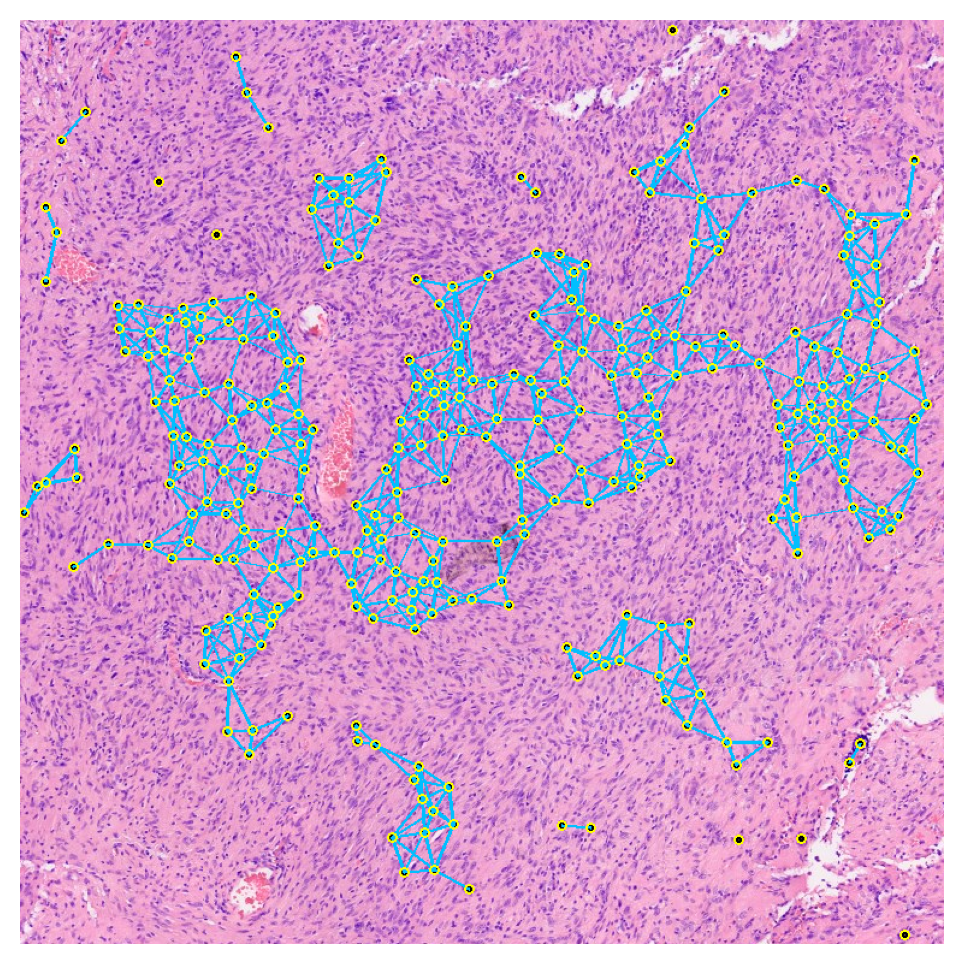} &
    \textbf{NGT} &
    LG-CNS &
    \textcolor{green!50!black}{\textbf{NGT}} &
    0.211 & 0.176 & 0.210 &
    Mesenchymal non-glial tumor with spindled architecture; overlaps with low-grade gliomas but distinguishable by growth pattern. \\
    \hline
    
    \rotatebox{90}{\texttt{Case\_4}} &
    \vcenterimg{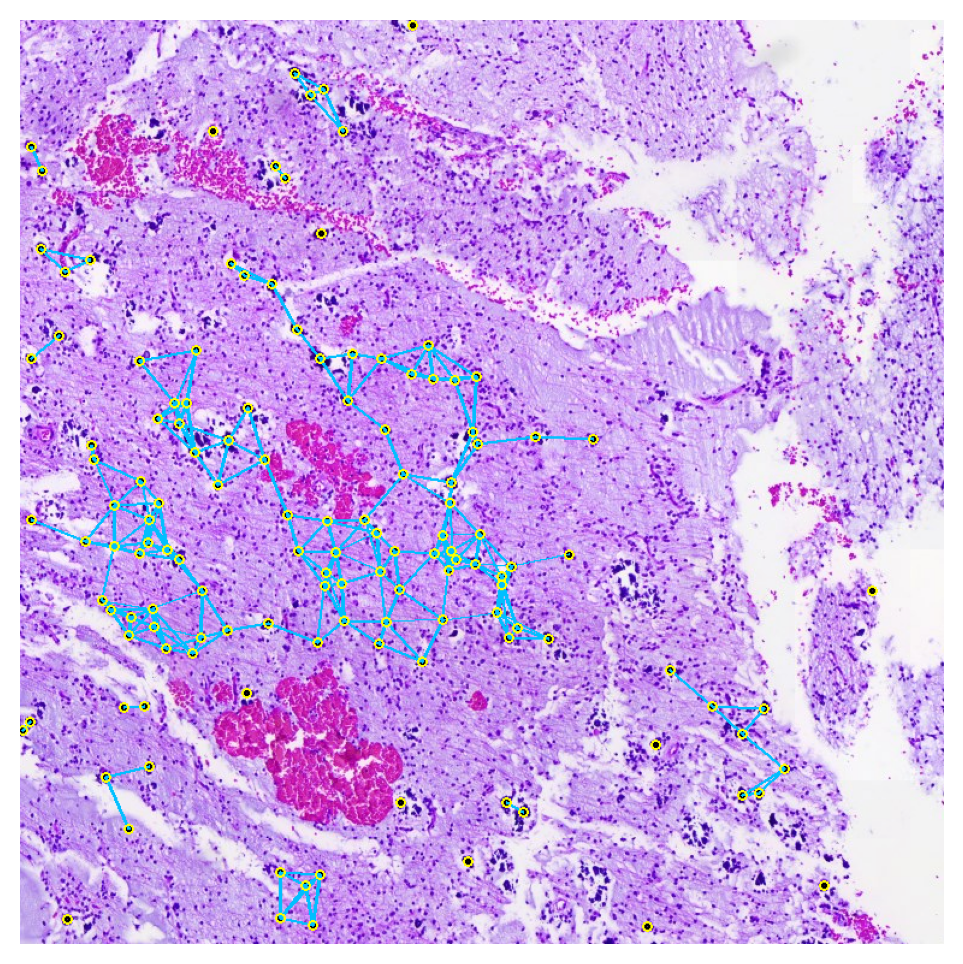} &
    \textbf{LG-CNS} &
    HG-CNS &
    \textcolor{green!50!black}{\textbf{LG-CNS}} &
    0.233 & 0.181 & 0.144 &
    Low-grade glioneuronal tumor with calcifications and mild nuclear density increase; lacks atypia, mitoses, or necrosis of high-grade tumors. \\
    \bottomrule
    
\end{tabular}
\label{tab:error-analysis}
\end{table*}

\noindent \textbf{Interpretability and Error Analysis.} Table~\ref{tab:error-analysis} details four cases where the image-only baseline fails to resolve grade (HG $\leftrightarrow$ LG) or lineage (HG $\leftrightarrow$ NG) confusions. PathMoE$_{WTG}$ correctly classifies these by leveraging text and nuclei graph modalities. The interaction weights reveal non-trivial graph contributions ($w_G$ ranging from 0.144 to 0.210), indicating that micro-architectural cues like nuclear density and spatial organization play a meaningful role in error correction. Clinical review by a neuropathologist emphasizes spatial and architectural features aligned with the graph modality. Text contributions are also appreciable ($w_t$ ranging from 0.167 to 0.184), suggesting that descriptive cues from pathology reports complement both visual and structural signals. While these examples are not exhaustive, they suggest that the MoE framework can dynamically leverage structural evidence when image-text signals are insufficient.


\noindent \textbf{Text Encoder Ablation Study.} To evaluate the impact of text representation quality, we swap the text encoder (TITAN, CONCH, BioMistral, BioBERT) while fixing all other components. TITAN consistently yields the strongest results across all PathMoE variants (Table \ref{tab:text_encoder_results}). In the EF$_{WTG}$ setting, TITAN achieves 0.799 macro-F1, surpassing BioMistral (0.776) and CONCH (0.685). This trend holds for the SG backbone (0.742 vs. 0.714 and 0.666). Notably, TITAN is built upon the CONCH framework and further optimized with WSI-oriented training, which likely yields more task-aligned representations for downstream multimodal fusion. While BioMistral is a much larger model (7B parameters), its lack of histopathology-specific pretraining limits its utility compared to TITAN. These findings emphasize that domain-aligned pretraining is more critical for multimodal fusion performance than model scale alone.

\begin{table*}[h!]
\centering
\caption{Macro precision (P), recall (R), and F1 for selected PathMoE variants under different text encoders.}
\label{tab:text_encoder_results}

\begingroup
\setlength{\tabcolsep}{2.0pt}
\renewcommand{\arraystretch}{1.06}
\setlength{\heavyrulewidth}{1.2pt}
\setlength{\lightrulewidth}{0.6pt}
\setlength{\cmidrulewidth}{0.45pt}

\fontsize{8pt}{9.6pt}\selectfont

\begin{adjustbox}{center}
\begin{tabular}{@{}l c c c c@{}}
\toprule
\textbf{Text encoder} & \textbf{Method} & \textbf{Macro P} & \textbf{Macro R} & \textbf{Macro F1} \\
\midrule

\multirow{2}{*}{\textbf{TITAN}}
& PathMoE-EF$_{\text{WTG}}$ & .812 & .801 & .799 \\
& PathMoE-SG$_{\text{WTG}}$ & .764 & .744 & .742 \\
\midrule

\multirow{2}{*}{\textbf{BioMistral}}
& PathMoE-EF$_{\text{WTG}}$ & .793 & .786 & .776 \\
& PathMoE-SG$_{\text{WTG}}$ & .741 & .717 & .714 \\
\midrule

\multirow{2}{*}{\textbf{BioBERT}}
& PathMoE-EF$_{\text{WTG}}$ & .725 & .702 & .699 \\
& PathMoE-SG$_{\text{WTG}}$ & .690 & .673 & .665 \\
\midrule

\multirow{2}{*}{\textbf{CONCH}}
& PathMoE-EF$_{\text{WTG}}$ & .708 & .693 & .685 \\
& PathMoE-SG$_{\text{WTG}}$ & .688 & .674 & .666 \\
\bottomrule
\end{tabular}
\end{adjustbox}

\endgroup
\end{table*}

\section{Conclusion}
PathMoE demonstrates that integrating foundation-model features with structured cell-graph domain knowledge via an interaction-aware multimodal architecture significantly improves pediatric brain tumor classification. By dynamically weighting unique, synergistic, and redundant modality interactions, our framework achieves superior performance while providing the sample-level interpretability essential for clinical trust in rare disease diagnosis.
%
%
%
\bibliographystyle{splncs04}
\bibliography{main}

\end{document}